\RequirePackage{fix-cm}
\documentclass[onecolumn,10pt]{svjour3}
\smartqed  
\usepackage{graphicx}
\usepackage{fixltx2e}
\usepackage[misc]{ifsym}
\usepackage{float}
\usepackage[multi-part-units=single]{siunitx}
\usepackage[font=small,labelfont=bf,tableposition=top]{caption}

\DeclareCaptionLabelFormat{andtable}{#1~#2  \&  \tablename~\thetable}
\begin{document}

\title{Force Estimation from OCT Volumes using 3D CNNs}



\author{Nils Gessert$^1$        \and
        Jens Beringhoff$^1$  \and
        Christoph Otte$^1$  \and
        Alexander Schlaefer$^1$  
}


\institute{\Letter \quad Nils Gessert, \email{nils.gessert@tuhh.de}, Tel.: +49 (0)40 42878 3389, https://orcid.org/0000-0001-6325-5092 \\ \\ $^1$ Hamburg University of Technology, Schwarzenbergstra\ss{}e 95, 21073 Hamburg}

\date{Preprint, published in IJCARS. DOI: }

\maketitle

\begin{abstract}

\textit{Purpose} Estimating the interaction forces of instruments and tissue is of interest, particularly to provide haptic feedback during robot assisted minimally invasive interventions. Different approaches based on external and integrated force sensors have been proposed. These are hampered by friction, sensor size, and sterilizability. We investigate a novel approach to estimate the force vector directly from optical coherence tomography image volumes.

\textit{Methods} We introduce a novel Siamese 3D CNN architecture. The network takes an undeformed reference volume and a deformed sample volume as an input and outputs the three components of the force vector. We employ a deep residual architecture with bottlenecks for increased efficiency. We compare the Siamese approach to methods using difference volumes and two-dimensional projections. Data was generated using a robotic setup to obtain ground truth force vectors for silicon tissue phantoms as well as porcine tissue.

\textit{Results} Our method achieves a mean average error of $\SI{7.7 \pm 4.3}{\milli\newton}$ when estimating the force vector. Our novel Siamese 3D CNN architecture outperforms single-path methods that achieve a mean average error of $\SI{11.59 \pm 6.7}{\milli\newton}$. Moreover, the use of volume data leads to significantly higher performance compared to processing only surface information which achieves a mean average error of $\SI{24.38 \pm 22.0}{\milli\newton}$. Based on the tissue dataset, our methods shows good generalization in between different subjects.

\textit{Conclusions} We propose a novel image-based force estimation method using optical coherence tomography. We illustrate that capturing the deformation of subsurface structures substantially improves force estimation. Our approach can provide accurate force estimates in surgical setups when using intraoperative optical coherence tomography.

\end{abstract}

\section{Introduction} \label{intro}

Robot-assisted minimally invasive surgery has become increasingly popular as it addresses various shortcomings of conventional minimally invasive surgery (MIS) \cite{Wilson.2014}. Robotic systems allow for motion scaling, tremor compensation and more degrees of freedom for tool movement which improves precision and reduces physical trauma \cite{Kroh.2015}. However, these systems often lack force feedback \cite{Diana.2015}, which would be helpful to control the instrument-tissue interaction during surgery. Typically, haptic feedback is generated on the patient side with haptic sensors, such as force sensors \cite{de2011force}. The information is fed back to a haptic interface that delivers the information to the human operator, e.g., as vibrotactile or kinesthetic feedback \cite{meli2017experimental}. One of the key challenges of generating reliable haptic feedback is accurate sensing of the forces at the patient  \cite{Okamura.2009haptic}. Lack of haptic feedback may lead to complications, increased completion time or severe injuries \cite{Pacchierotti.2015}. Although various approaches to realize force feedback have been proposed, the problem is still considered an open research challenge \cite{Bayle.2014}.

One approach is to directly incorporate force sensing devices into the robotic setup \cite{Puangmali.2012}. The devices can be placed inside or outside of the patient. If the device is placed outside the patient, e.g., in between tool and robot, only indirect measurement is possible. In addition to the forces at the tool tip, forces acting on the tool, e.g., due to friction, are measured which cannot be separated \cite{Faragasso.2014}. When placing the device closer to the tool-tissue interaction point, e.g., inside the tool tip, problems such as sterilization and biocompatibility arise \cite{Sokhanvar.2012}.

Due to these shortcomings, vision-based force estimation procedures have been proposed. First methods used a deformable template matching method to derive the force acting on an elstic object \cite{greminger2004vision}. Similar methods relying on mechanical deformation models have been studied for MIS scenarios \cite{Kim.2010haptic,noohi2014using,kim2012image}. Also, learning forces from image information using neural networks has been proposed \cite{greminger2003modeling}. More recent approaches have combined template matching and machine learning models \cite{Karimirad.2014,Mozaffari.2014}. Recently, recurrent neural networks (RNN) have been proposed to learn forces based on deformation tracked over time \cite{Aviles.2015,Aviles.2017towards}. The tissue surface is reconstructed from stereoscopic camera images and features representing surface deformation are defined. Then, the RNN is trained in a supervised fashion using ground-truth labels from a force sensor. Moreover, force estimation using optical coherence tomography (OCT) as an imaging modality has been proposed \cite{Otte.2016}. Surfaces are extracted from OCT volumes and forces are estimated based on surface deformation. 

So far, these approaches only rely on visual information capturing the surface deformation. Most proposed methods make use of stereoscopic cameras \cite{Rivero.2016} which are limited to the observation of surface deformations without imaging capabilities for inner tissue structures. Moreover, the proposed systems fuse features from visual deformation with robotic position feedback which limits the trained model to a specific robotic setup. Usually, retraining for a new setup is possible. However, quick adaptations might be difficult. 


We introduce a force estimation approach using volumetric OCT data and a novel Siamese 3D CNN architecture. The 3D CNN directly processes OCT volumes and outputs the three components of the related force vector. 

OCT can provide volumetric images with a resolution of a few micrometers which allows capturing the inner structure of tissue. In contrast to surface-based methods, the volumetric OCT image can also reflect tissue compression. Therefore, it is reasonable to expect OCT to provide a richer signal space with more information on subsurface deformation for accurate force estimation. We design our 3D CNN as a Siamese architecture that simultaneously processes the undeformed reference volume and the deformed sample volume to infer the force vector. 



In order to evaluate our method, we acquire data for a tissue phantom. We compare the approach to methods based on difference volumes. Moreover, we compare our method to surface-based force estimation approaches by only using surfaces extracted from OCT volumes with our Siamese CNN. 

Last, we validate our approach with a large dataset using porcine tissue. We use tissue from different subjects and vary the visible region for each tissue sample in order to show the robustness and generalization of our approach.

Subsequently, we describe our methodology in detail and then we provide and discuss our results. The results indicate that a precise force estimation is feasible. 

\section{Methods and Materials} \label{sec:methods}

\begin{figure*}
  \centering
  \includegraphics[width=1\textwidth]{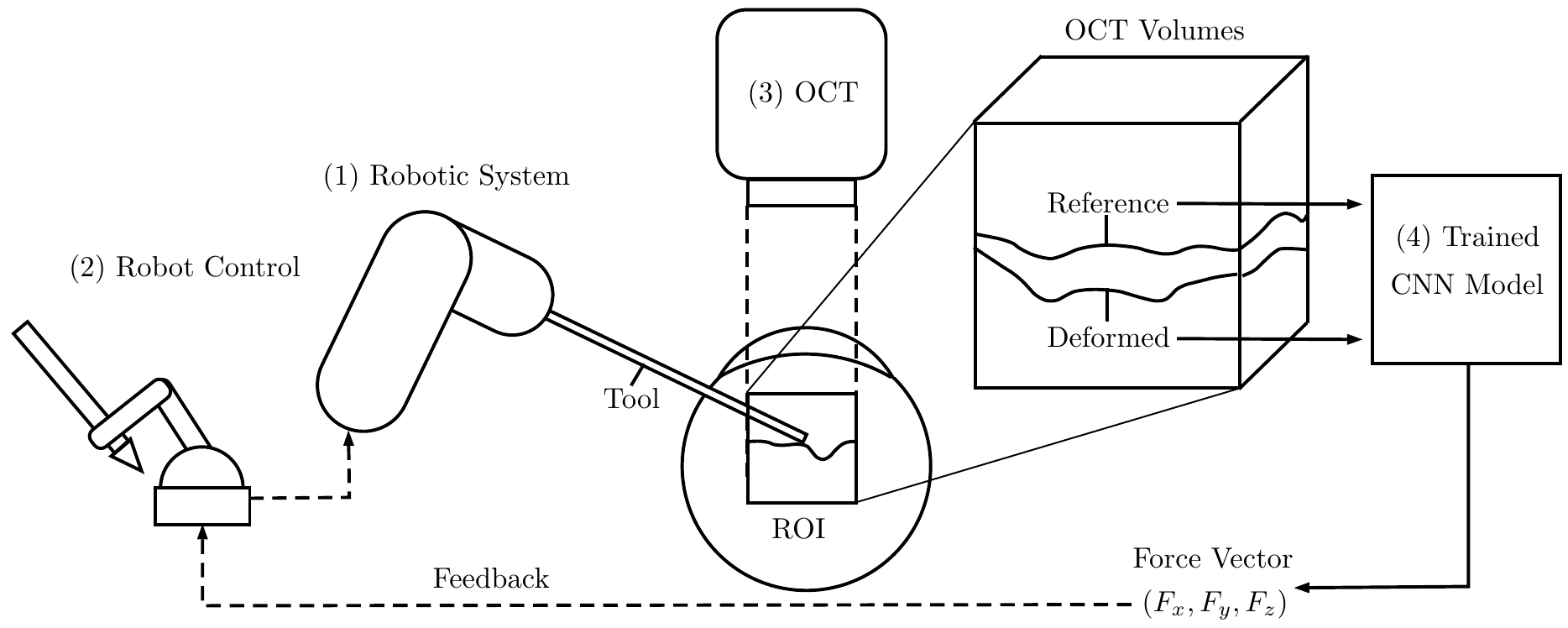}
\caption{The concept of our force estimation approach. A robotic system (1) that is controlled by a surgeon (2) performs actions in an ROI that lead to tissue-tool interactions. An intraoperative OCT device (3) repeatedly captures high resolution volumes of the ROI. The volumes are paired with a reference scan that are fed into a trained CNN (4) that performs inference in order to predict the force acting on the tissue. The force is fed back to the surgeon in order to provide visual or haptic feedback. }
\label{fig:general_setup}       
\end{figure*}


\subsection{Force Estimation with OCT}

The overall concept of using OCT image volumes to estimate the force during instrument tissue interaction is shown in Figure~\ref{fig:general_setup}. Surgery is performed with a robotic device that is remote-controlled by a surgeon. An OCT scan head is used to capture image volumes of the region of interest (ROI). First, a reference volume without tissue deformation is acquired. Then, when the tool deforms the tissue, more volumes are acquired. Both the reference and the current sample volume are fed into a trained CNN which predicts the force vector that acts on the tissue. Note, that the same vector with opposite direction acts on the tool and hence the predicted force is fed back to the surgeon and provides haptic or visual feedback. 
Force predictions are entirely image-based and independent of the robotic system performing the motion. Therefore, only the tool and the elastic tissue parameters are relevant for the predictions made by the CNN. As a result, models can be pre-trained for specific tools and tissue types.

\subsection{Experimental Setup} \label{sec:exp}

\begin{figure*}
  \centering
  \includegraphics[width=1\textwidth]{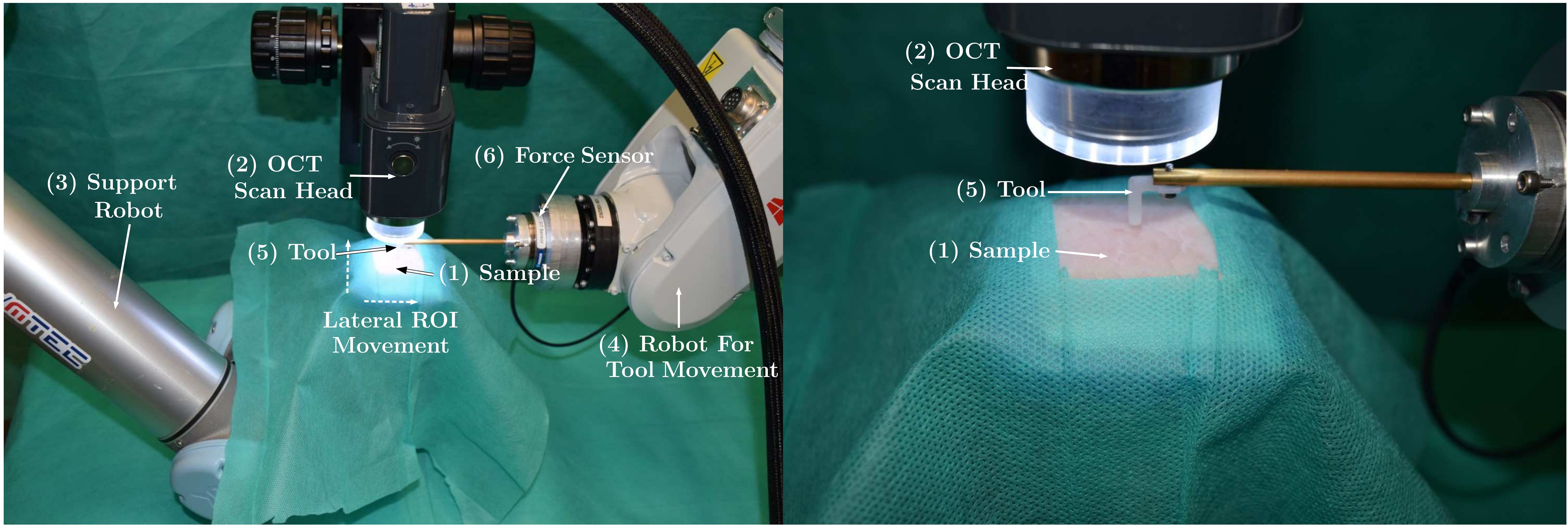}
\caption{The experimental setup we use for data acquisition. A tissue sample or phantom (1) is placed below the OCT device (2). A supporting robot (3) can move the sample in order to capture different ROIs during acquisition. The instrument robot (4) is equipped with a tool (5) and a force sensor (6) for ground-truth annotation of the OCT volumes. The main robot deforms the tissue with varying impact orientations.}
\label{fig:exp_setup}       
\end{figure*}

CNN training requires sufficiently large data sets. For automated data generation and systematic evaluation we use the setup shown in Figure~\ref{fig:exp_setup}. We place a phantom or tissue sample below the OCT scan head on a \textit{support} robot. The robot occasionally moves the sample in order to capture different ROIs. An \textit{instrument} robot is equipped with a tool and a force sensor at the tool base. Note, that the tool head is 3D printed and can be replaced. For each tool pose realized by the instrument robot we acquire an OCT image volume and the respective force vector. The data acquisition is performed as follows:

\begin{enumerate}
	\item Without deformation, the OCT device acquires a reference volume and the force sensor performs a reference measurement
	\item The instrument robot moves to a random orientation $\theta_x$, $\theta_y$, $\theta_z$ and deforms the sample with a random depth $d$
	\item With deformation, the OCT device acquires a sample volume and the force sensor performs a measurement
	\item After step (2) and (3) have been performed $L$ times, the support robot moves the sample in lateral directions by random values
	\item Steps (1) to (4) are repeated $M$ times
\end{enumerate}

As a result, for one iteration, we acquire $N=LM$ examples of reference and sample pairs with a force vector as a label for each pair. 

\subsection{OCT Imaging and Image Processing} \label{sec:oct}

\begin{figure*}
  \centering
  \includegraphics[width=1\textwidth]{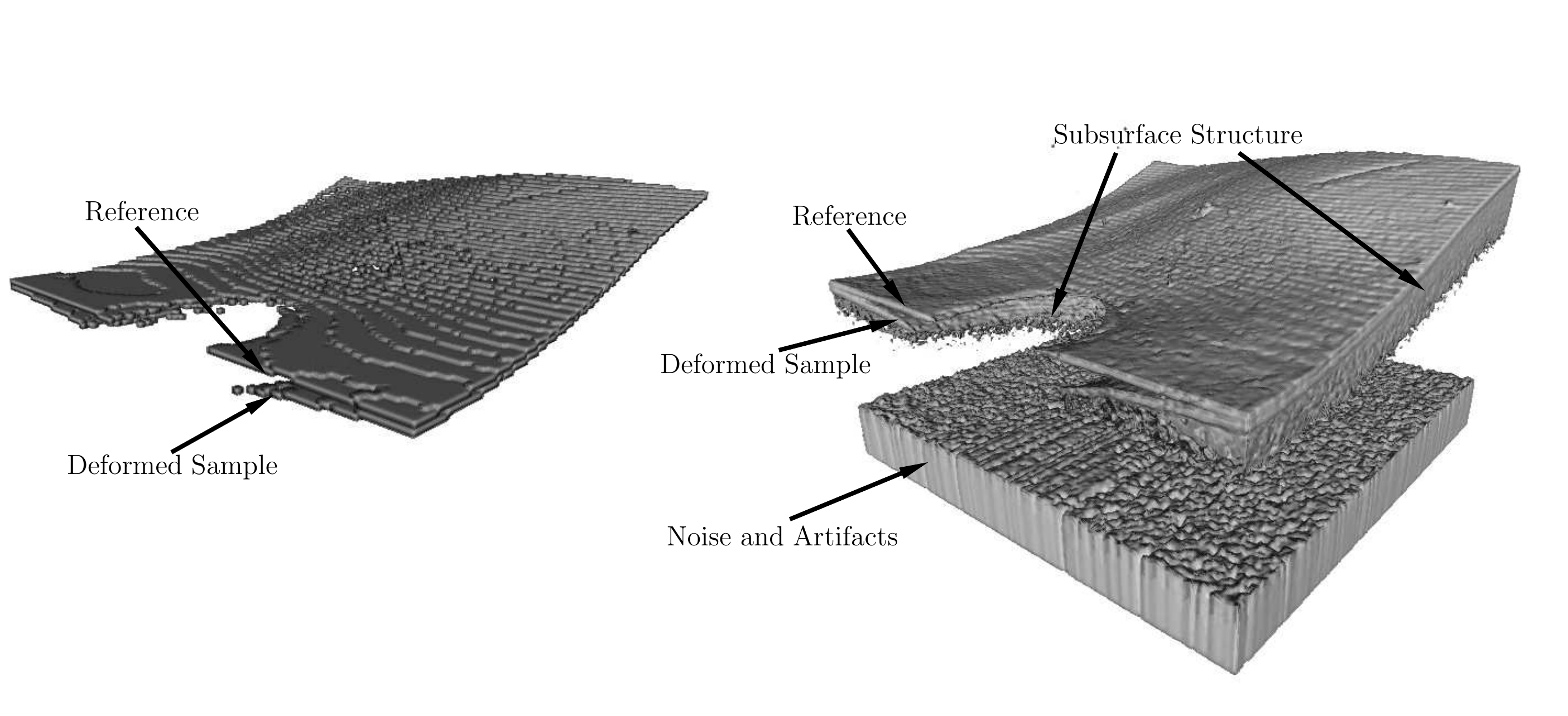}
\caption{Example data samples for training, shown as rendered volumes. On the right, full volumes are shown. On the left, the extracted surfaces are shown. In both cases, a deformed sample is overlayed with its reference measurement. A silicon phantom was used for these examples.}
\label{fig:vol_depth}       
\end{figure*}

The imaging device is a spectral domain OCT system which is based on interferometry. The method captures 1D depth profiles (A-Scans) using infrared light. Repeated scanning at neighboring lateral points results in a volume scan of the ROI. We use an OCT working at $\SI{1300}{\nano\metre}$ wavelength and therefore we can capture the inner structure of an ROI in up to $\SI{1}{\milli\metre}$ depth in scattering tissue. A single raw OCT volumes has a size of $128 \times 128 \times 512$ voxels. For our 3D CNN approach, we consider downsampled versions of size $64\times 64\times 64$ due to time, computational and memory limitations. 



We compare the volume-based method to approaches using the surface deformation only, extracted from the volumes. Maximum intensity projection (MIP) of the OCT volumes along the axial beam direction can be used for extraction \cite{Otte.2016}. The tissue phantom and tissue samples we use reflect the largest proportion of light at the surface. Therefore, the index at which the maximum intensity was observed represents a depth map of the tissue surface. Moreover, the intensity itself provides information of the surface characteristics as the intensity of the reflected light depends on the surface normal. We consider both as 2D surface representations for comparison to our volumetric approach.

Both the volumetric data and the depth images are shown in Figure~\ref{fig:vol_depth}. A deformed volume is overlayed with its corresponding reference volume. Note, that a shading of the tool is visible in the data which requires our model to be robust towards occlusion. As the occlusion is not related to the applied force, the 3D CNN can be expected to learn invariance towards the occlusion as CNNs have been shown to perform well at these kind of tasks \cite{Goodfellow.2016}.

\subsection{Model Definition and Training}

\subsubsection{Model Architecture} \label{sec:model}

\begin{figure*}[h]
  \centering
  \includegraphics[angle=90,width=1\textwidth]{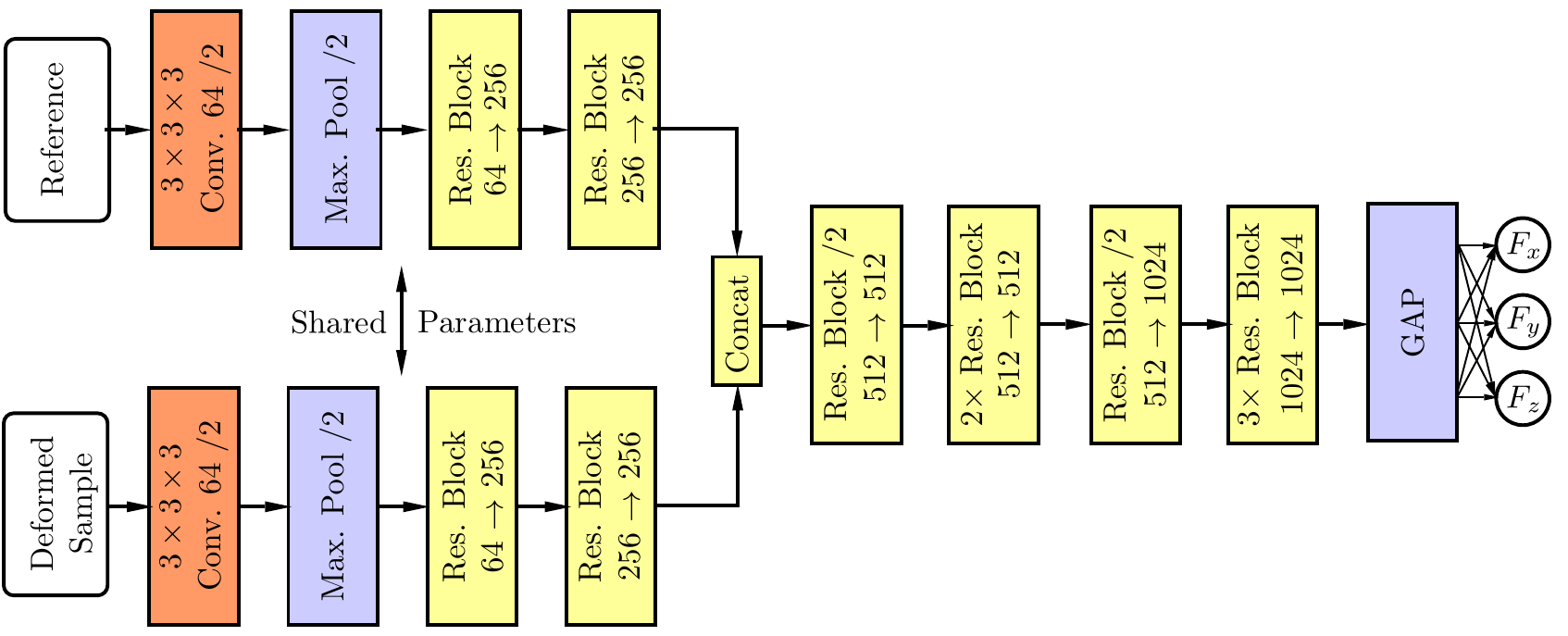}
\caption{The siamese 3D CNN architecture. The model takes a deformed sample and its corresponding reference volume as its input. In the initial part, the two volumes are processed independently up to a concatenation point. At this point, the feature maps are aggregated and processed jointly. At the output, global average pooling (GAP) is applied to the remaining feature maps and a fully-connected layer leads to the force vector output. \textit{Res. Block} refers to the residual blocks shown in Figure~\ref{fig:res_block}. $/2$ denotes a stride of two. Below each residual block, the change in the number of feature maps is denoted.}
\label{fig:cnn_arch}       
\end{figure*}

Our Siamese 3D CNN architecture is shown in Figure~\ref{fig:cnn_arch}. Siamese CNN architectures take two images to be compared as their input \cite{Leal.2016learning}. Then, the images are initially processed independently by the same set of learnable filters. At a concatenation point, the feature maps of both images are aggregated and processed jointly by the remaining network layers \cite{Zbontar.2015computing,Dosovitskiy.2015flownet}. 

\begin{figure}
  \centering
  \includegraphics[width=0.4\textwidth]{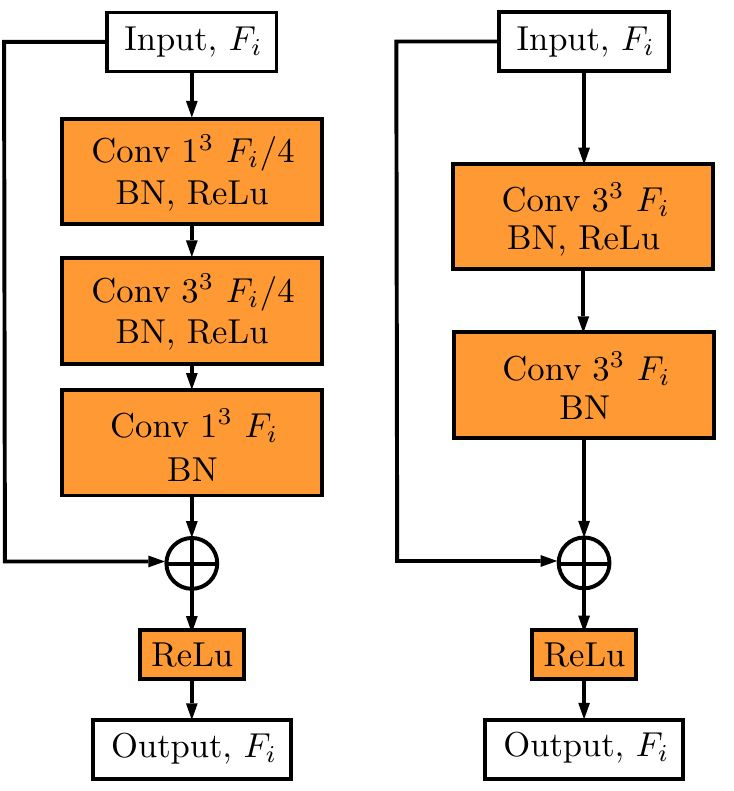}
\caption{The resdiual blocks for our architecture. $X^3$ denotes a $X\times X\times X$ filter. \textit{BN} denotes a batch normalization layer. $F_i$ denotes the number feature maps that this layer produces. Left, the residual block we employ with a bottleneck is shown. Right, a residual block without bottleneck is shown for comparison.}
\label{fig:res_block}       
\end{figure}

After the initial convolution in the network we employ residual blocks for an improved learning process \cite{He.2016b}. The blocks are shown in Figure~\ref{fig:res_block}. Furthermore, our residual blocks employ the \textit{bottleneck} concept \cite{He.2016}. Instead of directly applying convolutions, the input $x$ is downsampled first along its feature map dimension using a learnable $1\times 1\times 1$ filter. Then, the actual convolution with a larger $3\times 3\times 3$ filter is applied. Afterwards, the feature tensor is upsampled to its original feature map size. This method significantly reduces the number of learnable parameters and saves computational time. 

The concatenation of the two paths is defined as follows. For path 1, consider a tensor $t_1$ of shape $[N_B, W, H, D, F_1]$ where $N_B$ is the batch size, $W$, $H$, $D$, are the feature maps' width, height and depth and $F_1$ is the number of feature maps. Thus, the concatenated tensor $t_c = t_1 \| t_2$ has a shape of $[N_B, W, H, D, F_1+F_2]$. In our case, this doubles the number of feature maps which is why we keep the number of feature maps constant in the following spatial reduction ResNet block. This keeps the overall feature map sizes within the network at a reasonable level despite the concatenation.


Our general architecture choices are as follows. We use ReLu activation functions \cite{Glorot.2011}. Before every activation we use batch normalization in order to reduce internal covariate shift \cite{Ioffe.2015}. When we halve the spatial dimensions in our residual blocks, the $3\times 3\times 3$ filter uses a stride of two. We use nine residual blocks. From now on, we refer to our Siamese 3D CNN as SIAMCNN.



An alternative to a Siamese architecture is to use a single-path architecture that takes a difference or addition of volumes as its input. In this way, the deformation is captured in a single volume. We investigate performance when a single volume is passed to a 3D CNN that results from a subtraction or addition of the reference and deformed volume. The architecture is the same as the one shown in Figure~\ref{fig:cnn_arch}, except that one path is removed. We refer to this architecture as DIFFCNN\textsubscript{--} for subtraction and DIFFCNN\textsubscript{+} for addition.

Lastly, we introduced surface extraction with MIPs in Section~\ref{sec:oct}. We process the 2D depth representations with Siamese 2D CNN variants of our original architecture. The only difference to our model shown in Figure~\ref{fig:cnn_arch} is the use of 2D convolutions and 2D kernels instead of 3D convolutions and 3D kernels. We refer to this model as SURFCNN\textsubscript{MIP} for the 2D maximum intensity map as the model input and SURFCNN\textsubscript{DEPTH} for the 2D depth map as the model input.



\subsubsection{Training}

We train our models by minimizing the mean squared error (MSE) between ground-truth force labels and network predictions. We define the MSE as

\begin{equation}
	MSE = \frac{1}{d}\sum_{i=1}^{d}\frac{1}{N_B}\sum_{j=1}^{N_B}(y_i^{j}-\hat{y}_i^{j})^2
\end{equation}

where $d$ is the number of outputs, $N_B$ the batch size, $y$ the ground-truth label and $\hat{y}$ the network's predictions. 
We use the Adam algorithm \cite{Kingma.2014} for mini-batch gradient descent training. The initial learning rate is $l_r = \num{e-4}$. Every time the validation error plateaus, we divide the learning rate by a factor of 2 until no further improvement can be observed. As typically done for regression, we rescale the labels to a range of $[0,1]$ for training. 

We perform hyperparameter selection on the validation set with a grid search with limited bounds for relevant hyperparameters which include the number of residual blocks, the position of the concatenation point, the total number of feature maps and the learning rate schedule. Besides, we follow standard architecture design principles for filter and feature map size per layer \cite{Simonyan.2014}, batch normalization parameters \cite{Ioffe.2015} and Adam parameters \cite{Kingma.2014}.





\subsection{Materials and Datasets}

In our experimental setup we use a Thorlabs Telesto I SD-OCT device. Its lateral resolution is $\SI{15}{\micro\metre}$ and its depth resolution is $\SI{7.5}{\micro\metre}$. Its FOV covers a volume of $\SI{10 x 10 x 2.66}{\milli\metre}$ resulting in image volumes with a size of $\num{128 x 128 x 512}$ voxels. The instrument robot performing the deformations is an ABB IRB120 6-axis manipulator. The robot performs rotations of the tool tip with ranges of $[\SI{-30}{\degree},\SI{30}{\degree}]$, $[\SI{0}{\degree},\SI{10}{\degree}]$ and $[\SI{-10}{\degree},\SI{10}{\degree}]$ for $\theta_x$, $\theta_y$ and $\theta_z$, respectively. The deformation depth $d$ is in the interval $[\SI{0.5}{\milli\metre},\SI{1.5}{\milli\metre}]$. The support robot is a UR-5 6-axis manipulator. The force sensor for ground-truth annotation is an ATI Nano43. 

\begin{figure*}
  \centering
  \includegraphics[width=1\textwidth]{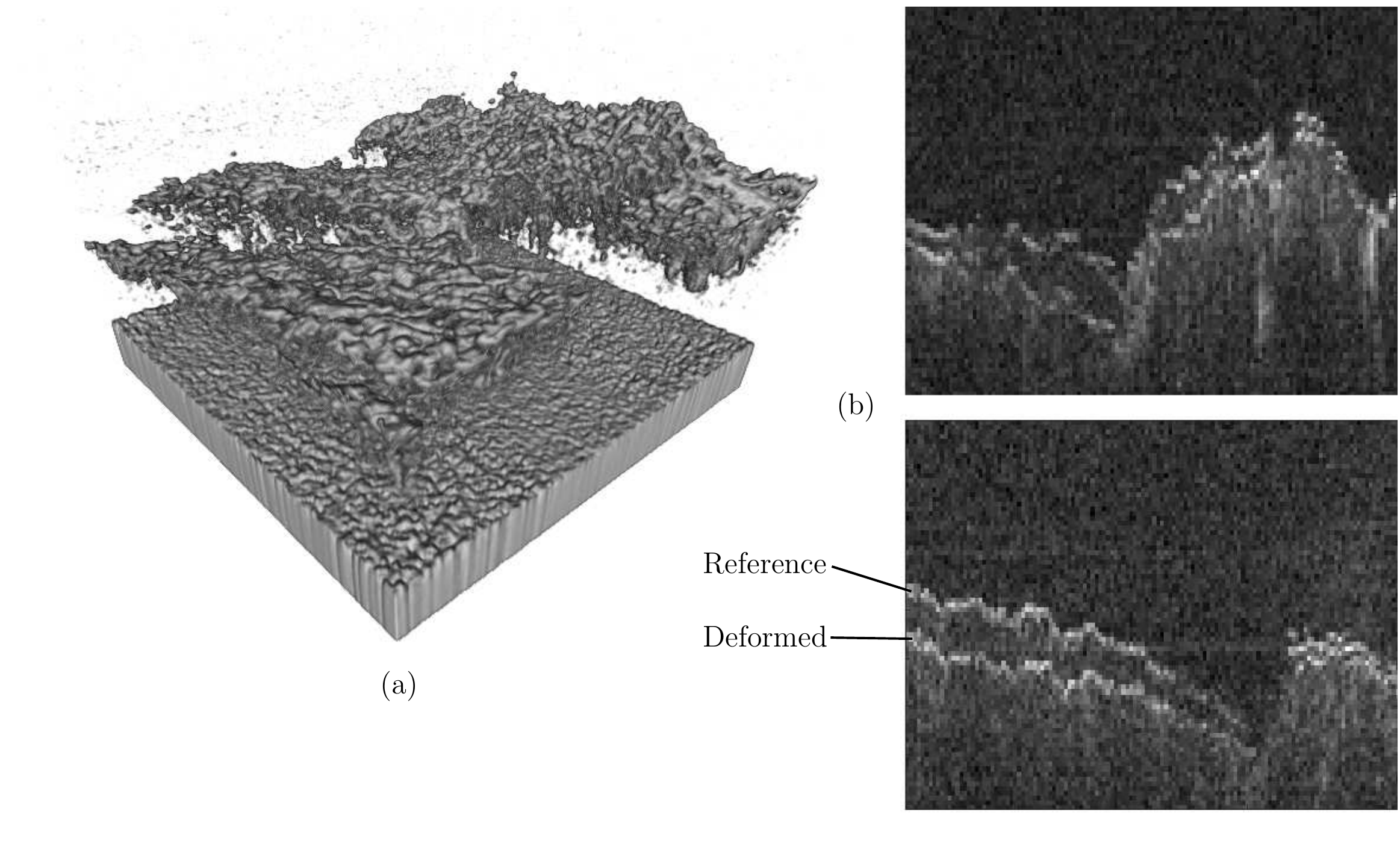}
\caption{Visualization of the tissue data. Left (a), a rendered volume of an OCT image that contains tissue is shown. Right (b), cropped, lateral slices through a volume are shown. The volume was created by overlaying a reference volume with a volume that contains deformed tissue.}
\label{fig:tissue}       
\end{figure*}

We acquired two datasets for the evaluation of our method. For the first dataset a silicon tissue phantom was used. In total, approximately $6600$ pairs of image volumes were acquired. We divide the set into training, validation and test set with ratios of $\SI{80}{\percent}$, $\SI{10}{\percent}$ and $\SI{10}{\percent}$. We fine-tune our models on the validation set and provide results for the test set. For the second dataset, porcine tissue was used. We acquired data with 17 different tissue samples from varying subjects. In total, approximately 8500 pairs of samples were acquired. We divide the set into training and test set with ratios of $\SI{80}{\percent}$ and $\SI{20}{\percent}$. For this dataset we do not use a validation set since we use the tuned models derived from the phantom dataset. An example volume is shown in Figure~\ref{fig:tissue}.

We use the TensorFlow Environment \cite{Abadi.2016} for implementation and train our models with an nVidia GTX 1080 Ti graphics card.

\subsection{Evaluation Strategy}

We use the mean average error (MAE) and average correlation coefficient (aCC) between network predictions and ground-truth labels for evaluation which are typical error metrics for regression \cite{Borchani.2015}. Furthermore, we show the per sample MAE error distribution with boxplots, both for the training and the test set. For relevant model variations we show the training times until convergence. 

\section{Results} \label{sec:results}

\begin{figure*}
  	\centering
  	\includegraphics[width=0.9\textwidth]{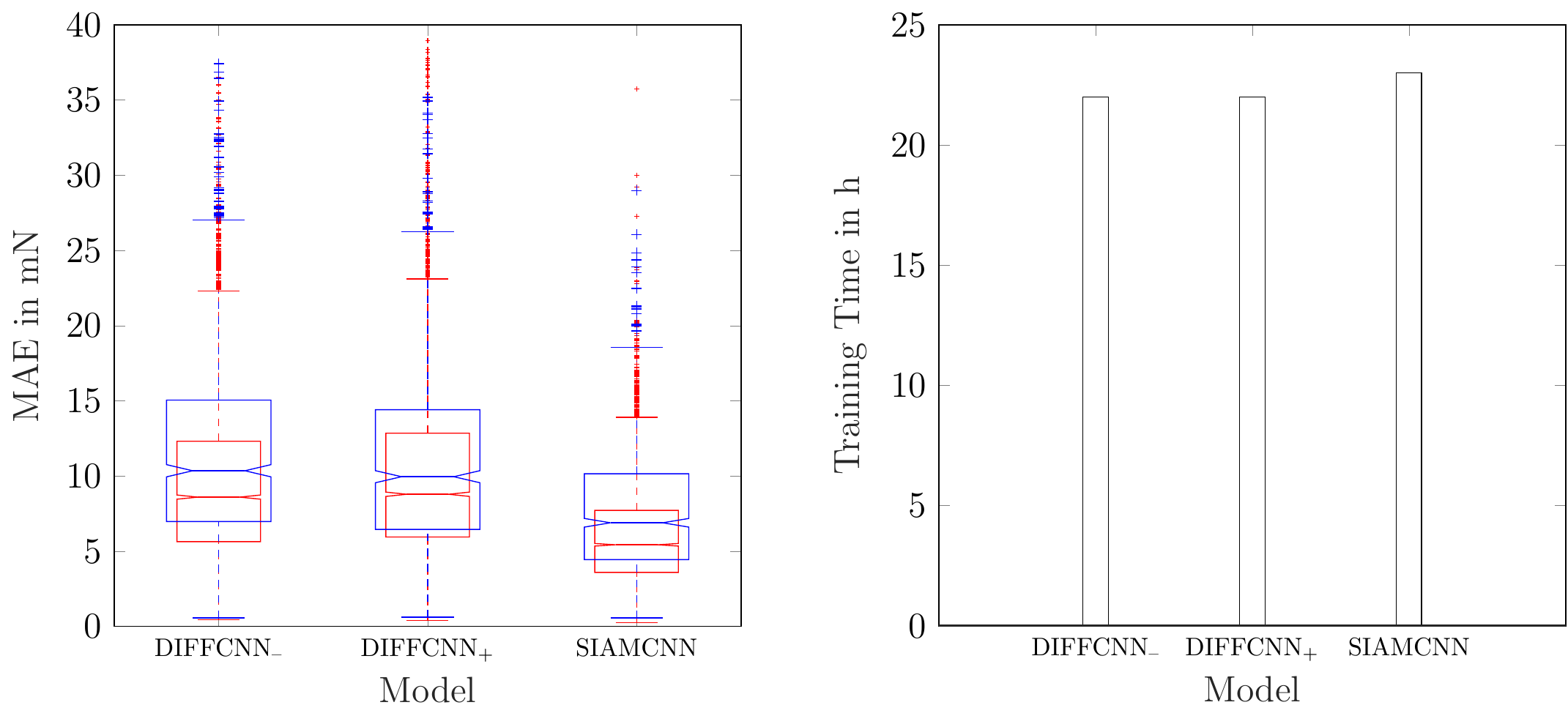}
	\\
	\vspace{10pt}
	\begin{tabular}{l l l l l}
		Model & DIFFCNN\textsubscript{--} & DIFFCNN\textsubscript{+} & \textbf{SIAMCNN} \\ \hline \\
		MAE & $\SI{11.76 \pm 6.9}{\milli\newton}$ & $\SI{11.59 \pm 6.7}{\milli\newton}$ & \boldmath $\SI{7.70 \pm 4.3}{\milli\newton}$ \\
		aCC & $0.973$ & $0.975$ & \boldmath $0.983$ \\ \hline \\
	
	\end{tabular} \\
    \captionlistentry[table]{A table beside a figure}
    \captionsetup{labelformat=andtable}
	\caption{Comparison for using the difference and addition of volumes compared to our Siamese approach. Top left, boxplots of the training MAE (red) and test MAE (blue) are shown. Top right, training durations until convergence are shown. Bottom, the MAE (with standard deviation) and the aCC are shown.}
	\label{fig:res_diff}
\end{figure*}

\begin{figure*}
  \centering
  \includegraphics[width=0.9\textwidth]{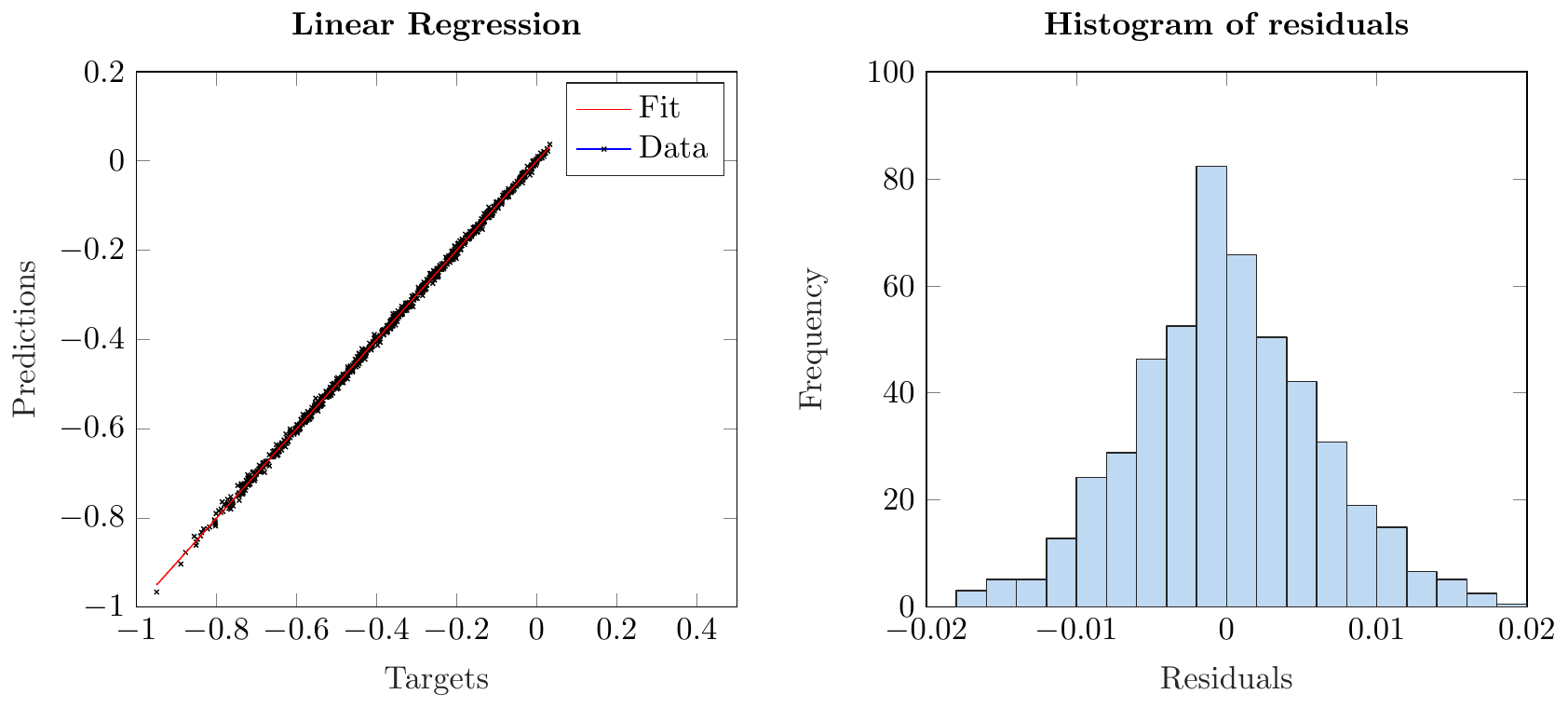}
\caption{Left, the linear regression plot between predictions and targets is shown. Right, the corresponding histogram of residuals is shown. Values are given in Newton. With an $R^2 = 0.99$ there is a strong relationship.}
\label{fig:regression}       
\end{figure*}



First, we compare SIAMCNN and DIFFCNN. The results are shown in Figure~\ref{fig:res_diff}. Generally, the aCC shows that the SIAMCNN model accurately learned force estimation. Joining the reference and sample volume with a subtraction or addition does not make a difference in terms of performance. SIAMCNN outperforms both single-path approaches while having a similar training time. Note, that real-time force estimation is feasible with an average processing time of $\SI{16.9 \pm 1.3}{\milli\second}$ for one instance of force estimation. 

\begin{figure*}
  	\centering
  	\includegraphics[width=0.9\textwidth]{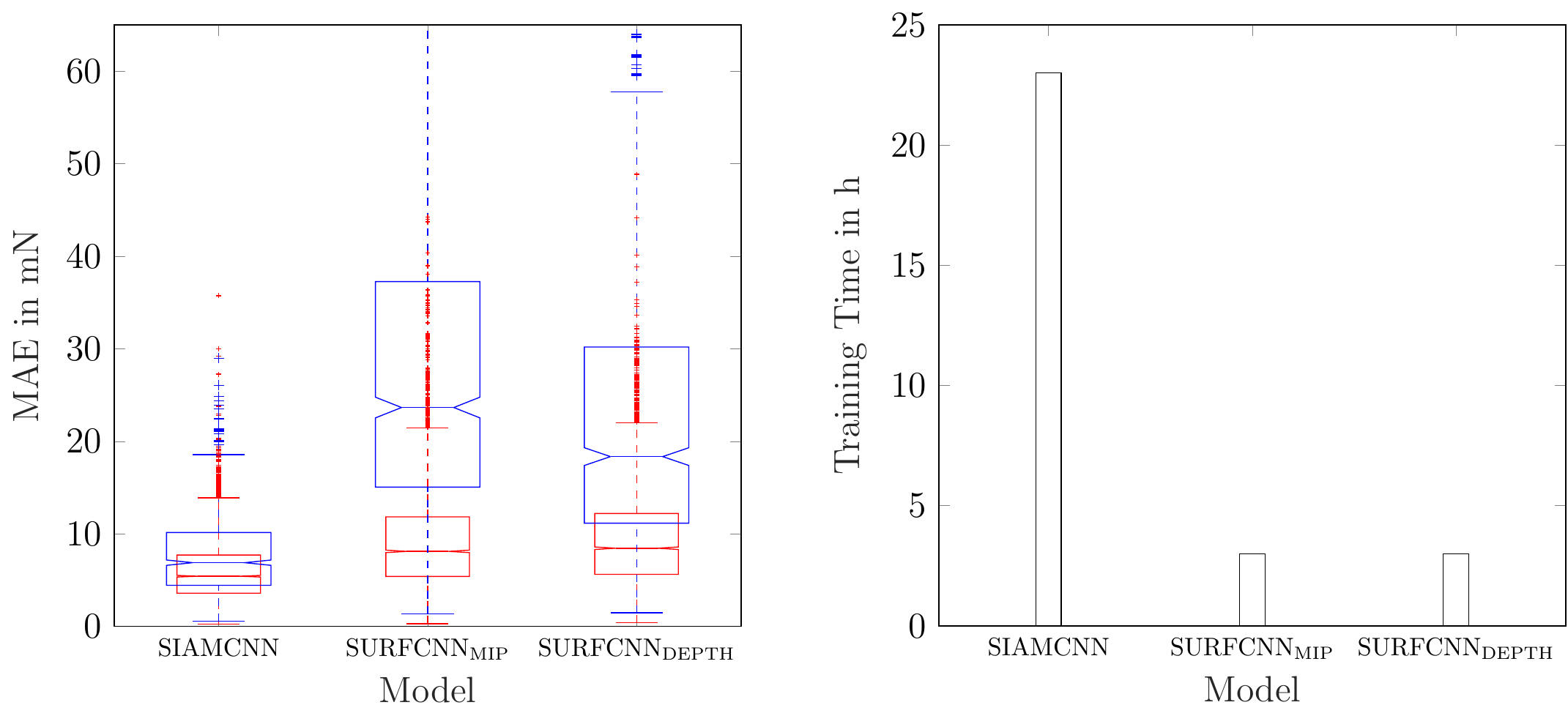}
	\\
	\vspace{10pt}
	\begin{tabular}{l l l l}
		Model & \textbf{SIAMCNN} & SURFCNN\textsubscript{MIP} & SURFCNN\textsubscript{DEPTH} \\ \hline \\
		MAE & \boldmath $\SI{7.70 \pm 4.3}{\milli\newton}$  & $\SI{29.42 \pm 22.9}{\milli\newton}$ & $\SI{24.38 \pm 22.0}{\milli\newton}$ \\
		aCC & \boldmath $0.983$ & $0.870$ & $0.877$ \\ \hline \\
	
	\end{tabular} \\	
    \captionlistentry[table]{A table beside a figure}
    \captionsetup{labelformat=andtable}
	\caption{Comparison of different 2D surface representations and volumetric inputs. Top left, boxplots of the training MAE (red) and test MAE (blue) are shown. Top right, training durations until convergence are shown. Bottom, the MAE (with standard deviation) and the aCC are shown. }
	\label{fig:res_2d}
\end{figure*}

Furthermore, we study the general properties of SIAMCNN. The regression plot in Figure~\ref{fig:regression} shows the linear relationship between model predictions and targets. There is a tight relationship with a high $R^2$ value of $0.99$. 

Furthermore, we compare our baseline model SIAMCNN to the surface-based 2D approaches SURFCNN\textsubscript{MIP} and SURFCNN\textsubscript{DEPTH}. The results are shown in Figure~\ref{fig:res_2d}. The volume-based model SIAMCNN significantly outperforms the two surface-based approaches. In terms of training time, the 2D models require substantially less training time.

\begin{figure}
  	\centering
  	\includegraphics[width=0.5\textwidth]{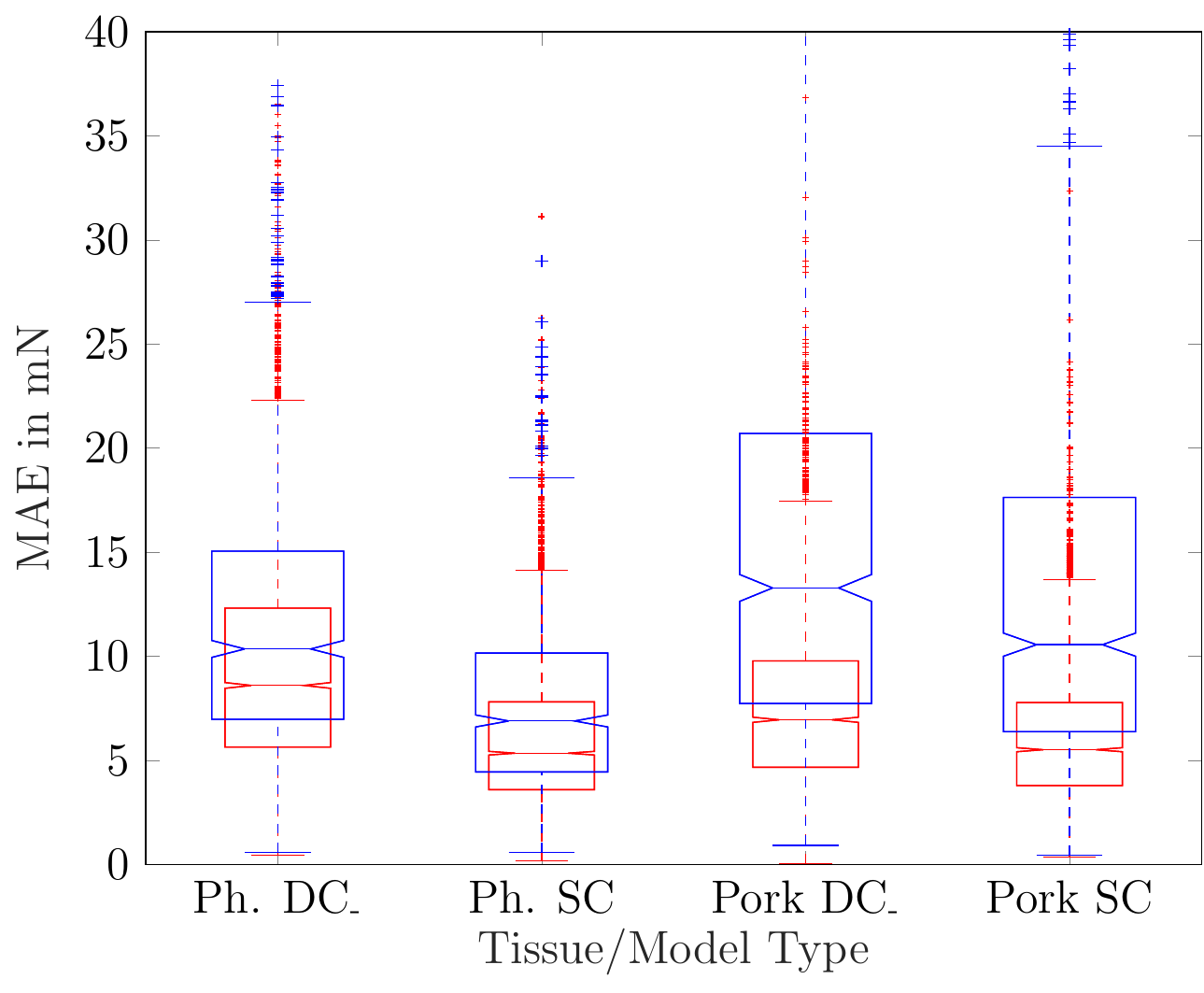}

	\caption{Comparison of phantom data to tissue. Boxplots of the training MAE (red) and test MAE (blue) are shown in $\si{\milli\newton}$. \textit{Ph.} refers to the phantom data. \textit{DC} refers to DIFFCNN and \textit{SC} refers to SIAMCNN.}
	\label{fig:res_pork}
\end{figure}

Lastly, we evaluate our method on a dataset with porcine tissue. The results are shown in Figure~\ref{fig:res_pork} and Table~\ref{tab:res_pork}. The errors are slightly larger than for the phantom data. All in all, the Siamese 3D CNN is able to generalize well to a new subject that was not present during model training. 

\begin{table}
	\centering
	\begin{tabular}{l l l}
	 & MAE & aCC \\ \hline \\
	Ph. DIFFCNN\textsubscript{--} & $11.76 \pm 6.9$ & $0.973$ \\
	Ph. SIAMCNN & $7.70 \pm 4.3$ & $0.983$ \\
	Pork DIFFCNN\textsubscript{--} & $14.21 \pm 8.2$ & $0.969$ \\
	Pork SIAMCNN & $11.42 \pm 5.9$ & $0.977$ \\ \hline \\
	\end{tabular}
	\caption{Comparison of phantom data to tissue. The MAE (with standard deviation) in $\si{\milli\newton}$ and the aCC are shown.}
	\label{tab:res_pork}
\end{table}

\section{Discussion} \label{sec:discussion}

We propose a novel method for image-based force estimation using OCT as an imaging modality. Image volumes are directly processed by a 3D CNN in order to predict a force acting between a tool and tissue. For this purpose, we introduce a novel Siamese 3D CNN architecture that processes a reference and a deformed sample simultaneously. 

The results in Figure~\ref{fig:res_diff} and Figure~\ref{fig:regression} show that our method accurately learned force estimation with an MAE of $\SI{7.70 \pm 4.3}{\milli\newton}$ and an aCC of $0.983$. This is achieved by only using two image volumes for one instance of force estimation. Many prior approaches depend on time series of deformations \cite{Rivero.2016,Carrasco.2016,Aviles.2016exploring} which is intractable for entire volumes to be processed by CNN models. Our model can be trained within one day on a standard consumer graphics card and allows for real-time force estimation.

Still, in some application scenarios, it might be difficult to obtain a reference volume, e.g., due to rapid tissue motion. Generally, the acquisition of reference volumes can be sped up by using a faster OCT system. Recently, swept source OCT systems with A-scan rates of multiple MHz have been demonstrated and commercial systems with 1.5 MHz A-scan rate are available \cite{potsaid2010ultrahigh}. Moreover, full-filed OCT systems for fast parallel volume acquisition have been proposed \cite{hillmann2013off}. Using such systems, small tissue patches at an instrument tip can be imaged with several hundred volumes per second, i.e., motion artifacts and latency would be minimal. Clearly, our method could be readily applied to OCT image volumes from such OCT devices.

Besides SIAMCNN, we consider a subtraction or addition of the reference and deformed volumes allowing for a single-path 3D CNN, DIFFCNN. This relates to previous approaches where differences of surfaces were considered for force estimation \cite{Otte.2016}. SIAMCNN is more accurate with an error of $\SI{7.70 \pm 4.3}{\milli\newton}$ compared to $\SI{11.76 \pm 6.9}{\milli\newton}$ for DIFFCNN\textsubscript{--}. The improved performance implies that learning distinct preprocessing for both volumes within the 3D CNN is beneficial for force estimation. At the same time, there is hardly any difference in terms of training duration as the SIAMCNN and DIFFCNN models contain almost the same number of parameters. Therefore, our SIAMCNN model improves performance without demanding more resources.



Prior approaches relied on deformations that were obtained from surface reconstructions \cite{Aviles.2015} or surface extraction \cite{Otte.2016}. For SIAMCNN, volume processing is superior with an error of $\SI{7.70 \pm 4.3}{\milli\newton}$ compared to $\SI{24.38 \pm 22.0}{\milli\newton}$ for the best performing 2D model SURFCNN\textsubscript{DEPTH}. This suggests that capturing subsurface tissue compression with OCT image volumes allows for learning richer feature representations. Furthermore, the training and test error distribution depicted in Figure~\ref{fig:res_2d} show that overfitting occurs for the 2D case, despite our early-stopping training strategy. This also indicates that the surfaces extracted from OCT volumes carry fewer generalizable features than volumes.

In a last step, we validate our results on an animal tissue dataset that contains samples from different ROIs and varying subjects. For deep learning methods, overfitting is often an issue. In our case, the 3D CNN might overfit to subject-specific features which would hinder application in practice where the model is always applied to a new subject. Therefore, it is important to ensure that the model actually learned tissue- and not subject-specific deformation features. Our test set for the tissue datasets contains samples from a subject that was not present during training. The test set results in Figure~\ref{fig:res_pork} show that our model was able to produce accurate results despite subject variations. Therefore, similar to previous force estimation approaches \cite{Aviles.2017towards}, our method is able to achieve invariance towards small subject variations.

As our method directly learns relevant features from the volume data, the models do not require any adjustments for other tissue types and can be directly trained on new datasets. As a drawback, this requires acquisition of new datasets for new tissue types. With intraoperative OCT systems spreading in availability \cite{ehlers2014integrative}, our method could see application in practice, given extended validation with human operators who receive the estimated forces as feedback.

\section{Conclusions} \label{sec:conclusion}

We address force estimation for tool-tissue interaction in surgery. For sensorless measurement, we propose a novel image-based force estimation method using OCT volume data. We associate a reference volume measurement with a volume of deformed tissue for force prediction with a 3D CNN. In order to process both volumes we introduce a novel Siamese 3D CNN architecture. OCT allows to capture inner structure of tissue. We demonstrate that exploiting deep tissue structures with volumes performs significantly better than deriving deformation from the surface in OCT volumes. Lastly, we show the applicability of our method on a tissue dataset where generalization to new subjects that were not present during training is feasible.

\section*{Conflict of Interest}

The authors declare that they have no conflict of interest.

\section*{Ethical Approval}

This article does not contain any studies with human participants or animals performed by any of the authors.

\section*{Informed Consent}

Informed consent was obtained from all individual participants included in the study.


\bibliographystyle{spmpsci} 
\bibliography{egbib}   

\end{document}